\documentclass[letterpaper, 10 pt, conference]{ieeeconf} 
\IEEEoverridecommandlockouts
\newcommand\copyrighttext{%
	\footnotesize This work has been submitted to the IEEE for possible publication. Copyright may be transferred without notice, after which this version may no longer be accessible.
}
\newcommand\copyrightnotice{%
	\begin{tikzpicture}[remember picture,overlay]
	\node[anchor=south,yshift=10pt, xshift=10pt] at (current page.south) {\fbox{\parbox{\dimexpr\textwidth-\fboxsep-\fboxrule\relax}{\copyrighttext}}};
	\end{tikzpicture}%
}

\usepackage{printlen}

\usepackage[T1]{fontenc}
\usepackage{amsmath,amsfonts}
\usepackage{cite}
\usepackage{stfloats}
\usepackage{graphicx}
\usepackage[caption=false,font=scriptsize]{subfig}
\usepackage{url}
\usepackage{hyperref}

\usepackage[per-mode = fraction]{siunitx}
\usepackage{acronym}
\usepackage[none]{hyphenat}

\usepackage{bbm} 

\usepackage{booktabs}
\usepackage{tabularx}
\usepackage{multirow}
\usepackage{pifont}

\usepackage{tikz, pgfplots, pgfplotstable}
\usetikzlibrary{arrows, arrows.meta, patterns}
\usepgfplotslibrary{statistics}
\pgfplotsset{compat=1.18}

\newcommand{\coloredbox}[1]{
  \tikz{
    \draw[fill=#1] (0,0.0) rectangle (0.5,0.2);
  }
}

\usepackage{xcolor}
\definecolor{Black}{HTML}{000000}
\definecolor{Blue}{HTML}{0065bd}
\definecolor{Bluelight}{HTML}{D6E8F7}
\definecolor{Bluestrong}{HTML}{003359}
\definecolor{Red}{HTML}{8C000F}
\definecolor{Orange}{HTML}{E37222}
\definecolor{OrangePP}{HTML}{E97132}
\definecolor{Green}{HTML}{A2AD00}
\definecolor{GreenCR}{HTML}{008000}
\definecolor{LightGray}{HTML}{e7e7e7}
\definecolor{Gray}{HTML}{7f7f7f}
\definecolor{Gray-opac}{HTML}{d8d8d8}

\graphicspath{{./figures/}}

\title{\LARGE \bf
Learning to Sample: Reinforcement Learning-Guided Sampling for Autonomous Vehicle Motion Planning
}

\author{Korbinian Moller, Roland Stroop, Mattia Piccinini, Alexander Langmann, Johannes Betz%
\thanks{K. Moller, R. Stroop, M. Piccinini, A. Langmann and J. Betz are with the Professorship of Autonomous Vehicle Systems, TUM School of Engineering and Design, Technical University of Munich, 85748 Garching, Germany; Munich Institute of Robotics and Machine Intelligence (MIRMI).}%
}

\begin{document}
\bstctlcite{BSTcontrol}

\maketitle
\copyrightnotice


\begin{abstract}
Sampling-based motion planning is a well-established approach in autonomous driving, valued for its modularity and analytical tractability. In complex urban scenarios, however, uniform or heuristic sampling often produces many infeasible or irrelevant trajectories. We address this limitation with a hybrid framework that learns where to sample while keeping trajectory generation and evaluation fully analytical and verifiable. A reinforcement learning (RL) agent guides the sampling process toward regions of the action space likely to yield feasible trajectories, while evaluation and final selection remains governed by deterministic feasibility checks and cost functions. We couple the RL sampler with a world model (WM) based on a decodable deep set encoder, enabling both variable numbers of traffic participants and reconstructable latent representations. The approach is evaluated in the CommonRoad (CR) simulation environment and compared against uniform-sampling baselines, showing up to \SI{99}{\percent} fewer required samples and a runtime reduction of up to \SI{84}{\percent} while maintaining planning quality in terms of success and collision-free rates. These improvements lead to faster, more reliable decision-making for autonomous vehicles in urban environments.

\end{abstract}

\vspace{0.1cm}



\section{Introduction}
\label{sec:introduction}

Autonomous vehicles are expected to operate safely and efficiently in dense, unpredictable environments. Urban scenarios are particularly challenging due to constrained spaces, complex traffic dynamics, and the need for rapid yet reliable decision-making. In this context, a motion planner must generate feasible trajectories under dynamic constraints while minimizing computational delays. Sampling-based motion planning has become a widely adopted approach because of its modularity, analytical tractability, and compatibility with safety frameworks~\cite{Werling2010, Frenetix}. However, uniform or heuristic-based sampling often proves inefficient, as many sampled goal states lead to trajectories that are either infeasible or pass through obstacle-prone regions~\cite{Frenetix}. This inefficiency directly impacts responsiveness in time-critical scenarios. End-to-end learning approaches attempt to mitigate these issues by directly predicting trajectories, but they typically sacrifice interpretability and verifiability, limiting their applicability in safety-critical systems~\cite{Aradi2022, Dauner2023CORL}. Hybrid methods that combine learning with analytically grounded modules offer a promising balance between adaptability and reliability~\cite{Ichter2018, Dauner2023CORL}. Building on this idea, we propose a modular motion planning framework in which an RL agent guides the selection of goal states, while trajectory generation and evaluation remain analytical and verifiable. By biasing sampling toward states more likely to produce feasible and context-appropriate trajectories, our approach addresses the inefficiency of uniform sampling, as shown in \autoref{fig:introduction}. To incorporate structured scene understanding, the RL module is coupled with a WM~\cite{Hafner2025} based on a decodable deep set encoder~\cite{Kortvelesy2023}. This design enables reasoning over variable numbers of surrounding agents and allows reconstructable latent representations. In summary, this paper makes the following contributions:
\begin{itemize}
\item We introduce a hybrid motion planning framework that combines RL for goal state sampling with analytical and safety-verifiable trajectory generation and evaluation.  
\item We release a WM framework for road traffic with variable-sized traffic and road representations based on the integration of deep sets~\cite{Kortvelesy2023} into Dreamer V3~\cite{Hafner2025}.
\item We evaluate the framework in simulation, showing that, compared to the uniform sampling baseline, it maintains planning quality while reducing the number of required samples by up to \SI{99}{\percent} and runtime by up to \SI{84}{\percent}.
\end{itemize}

\input{figures/introduction_v2.tex}

\section{Related Work}
\label{sec:relatedwork}

Motion planning methods in autonomous driving are commonly divided into optimization-, graph-, sampling-, and learning-based approaches~\cite{Gonzalez2015, Paden2016, Teng2023, Zhou2022, Dong2023}. In sampling-based motion planning (SBMP), trajectories are generated by connecting sampled goal states to the vehicle’s current state and evaluated with cost functions such as travel time, comfort, or collision probability~\cite{Frenetix, Moller2025}. This modularity makes SBMP particularly appealing for autonomous driving, as cost and feasibility layers can be flexibly adapted without redesigning the entire algorithm.

A key limitation of SBMP lies in their reliance on uniform or heuristic sampling, which produces many infeasible trajectories and inflates runtime in dense traffic~\cite{Frenetix}. To mitigate this inefficiency, analytical and learning-based refinements have been proposed.

\subsection{Analytical Motion Planning Enhancements}

One major direction is to restrict the sampling domain via reachable sets. Early work computed conservative envelopes of other traffic participants~\cite{Althoff2016}, while later methods extracted collision-free intervals in the Frenet frame to reduce sampling and generate jerk-optimal trajectories~\cite{Wursching2021}. Recent extensions integrate reachable sets with iterative LQR optimization to balance feasibility with smoothness~\cite{Liu2024}. Corridor-based approaches constrain planning to rule- and geometry-compliant regions~\cite{IraniLiu2023}, ensuring safety but limiting flexibility.

Beyond structural restrictions, adaptive and focused sampling methods bias exploration toward relevant regions. Adaptive densities increase around critical obstacles while remaining sparse elsewhere~\cite{Li2020}. Another accelerates planning in dense traffic by combining path–velocity decomposition with binary-search–based collision checking~\cite{Zhu2023}. Lane-centered informed sampling with reuse of prior solutions has also been proposed to improve comfort and efficiency in urban driving~\cite{Smit2022}. A coarse-to-fine refinement strategy further narrows computation to promising areas of the state space~\cite{Sun2023}. In robotics, non-uniform RRT* with convex partitioning achieves similar gains but lacks semantic traffic information~\cite{Wilson2021}.

\subsection{Learning-Based Motion Planning Approaches}

Beyond analytical refinements, learning-based approaches have been explored to increase adaptability in motion planning. End-to-end approaches directly map sensors to controls but lack interpretability~\cite{Aradi2022, Scheel2021}, limiting safety-critical deployment. More promising are modular methods where learning augments classical pipelines.

Learning-guided sampling leverages freespace forecasting~\cite{Hu2021}, learned non-uniform distributions~\cite{Ichter2018}, or imitation learning combined with prediction~\cite{Zhang2020,Zhang2022}. RL and inverse RL have also been integrated: offline RL optimizes highway planning~\cite{Mirchevska2023}, IRL injects human preferences into cost functions~\cite{Trauth2023}, and online RL adapts weights in real time~\cite{Trauth2024}. Hybrid safety filters such as SafetyNet~\cite{Vitelli2022} and Frenet-based networks like LF-Net~\cite{Yu2024} demonstrate the trade-off between context awareness and verifiability.
RL approaches have also been applied in racing scenarios, where interaction is highly dynamic. Early work employed curriculum RL for progressive overtaking~\cite{Song2021}, followed by demonstrations that deep RL can outperform human drivers in high-fidelity simulations~\cite{Wurman2022}. More recent efforts integrate safety fallbacks into RL-based overtaking planners~\cite{Ogretmen2024}. While these works demonstrate RL’s strengths in adversarial contexts, they generalize poorly to structured urban traffic.

Finally, WMs provide a foundation for bridging learning and classical planning. Model-based RL improves sample efficiency by reusing interaction data within a learned dynamics model~\cite{Ha2018, Sekar2020}. Recent advances include Dreamer V3~\cite{Hafner2025}, which achieves state-of-the-art control via imagined rollouts. In autonomous driving, surveys emphasize their potential for prediction and planning~\cite{Guan2025}, while adaptive WMs integrated with MPC achieve top performance in simulation with improved generalization~\cite{Vasudevan2025}. Generative approaches such as DriveDreamer learn predictive models from real driving videos for controllable simulation and policy training~\cite{Wang2024}.

In summary, learning-based methods enhance adaptability but often compromise interpretability and safety, motivating our hybrid framework that preserves analytical trajectory generation and evaluation.

\section{Methodology}
\label{sec:method}

\autoref{fig:method_overview} provides a high-level overview of our hybrid motion planning framework. Time-varying observations $\mathcal{O}_t$ are encoded by a learned WM into a model state $s_t$ (\autoref{subsec:worldModel}). An RL agent operates on $s_t$ and proposes a promising goal specification \(g(\tau)\) (\autoref{subsec:reinforcementLearningAgent}) in a curvilinear reference frame $(s,d)$ around a given reference path~$\Gamma$~\cite{Wursching2024}. The reference path is assumed to be provided by an upstream global planner and extends from the initial ego position to the global target region. Depending on the setting, additional goals may be sampled around the proposal \(g(\tau)\). Each goal is then connected to the initial state $x(0)$ with a continuous-time trajectory candidate~$\xi$ by a deterministic motion planner~\cite{Frenetix}. The resulting candidates form the trajectory set~$\mathcal{T}$. The optimal trajectory is selected by minimizing a weighted sum of costs over $\mathcal{T}$ (\autoref{subsec:deterministic_planner}).

\begin{figure*}[!t]
    \centering
    \input{figures/methodology.tex}%

    \vspace{-5mm}
    \caption{High-level framework. The WM encodes the environment into a latent state $z_t$, where an RL agent selects high-level end-conditions $g(\tau)$. A deterministic planner then generates and evaluates trajectories that satisfy kinematic and safety constraints before producing the optimal trajectory. Blue paths in the diagram indicate components that are only used during training.}
    \label{fig:method_overview}
\end{figure*}

\subsection{World Model}
\label{subsec:worldModel}

WMs learn predictive latent representations of an environment's dynamics. In the context of RL, WMs enable imagination-based rollouts that improve sample efficiency and transferability compared to purely model-free approaches. Model-free methods typically require large amounts of interaction data, struggle to generalize across unseen situations, and offer limited interpretability. Model-based methods mitigate these drawbacks by learning dynamics that can be reused for training, evaluation, and fast adoption across tasks. This is particularly relevant in autonomous driving, where collecting data is costly and the ability to adapt to new road geometries and varying numbers of traffic participants is crucial. 
%
We build on DreamerV3~\cite{Hafner2025}, which learns latent dynamics from replayed trajectories and trains the actor $\pi_\theta$ and critic $V_\phi$ networks using imagined rollouts. In the following, we focus on the WM, while actor and critic training is described in the next subsection. The WM is modeled as a Recurrent State-Space Model (RSSM) that consists of an encoder, a recurrent dynamics model, and a decoder. At each timestep, the encoder maps structured observations $\mathcal{O}_t$ combined with the recurrent state $h_t$ to stochastic latent representations $z_t$. 
These jointly constitute the WM state $s_t \doteq \{h_t, z_t\}$.
Based on the model states and sampled actions, the WM imagines future latent trajectories and predicts rewards that serve as input to the actor–critic
learning process.
The decoder reconstructs the structured observations from model states to guide the WM in learning meaningful representations.

Unlike prior applications of Dreamer with image inputs, we use high-level features. This avoids learning perception from scratch and does not assume a bird’s-eye view. The observation vector consists of ego-centric state variables $\bar{o}^{\mathrm{EC}}_t$, a set of obstacles $\hat{o}^{\mathrm{obs}}_t$, a set of lanelet features $\hat{o}^{\mathrm{lane}}_t$, and the environment reward $r_t$. Sets are required because the number of obstacles and lanelets varies between scenarios, and their ordering carries no additional semantic meaning. We factorize the observation as:
\begin{equation}
\mathcal{O}_t = \bigl[\bar{o}^\mathrm{EC}_t,\; \hat{o}^{\mathrm{obs}}_t,\; \hat{o}^{\mathrm{lane}}_t,\; r_t\bigr].
\label{eq:obs_factorization}
\end{equation}
\autoref{fig:obs_space} illustrates how observations are mapped from a traffic scenario into this structured representation.
\begin{figure}[!ht]
  \centering
    \input{figures/observation.tex}%

  \vspace{-7mm}
  \caption{A driving scenario is mapped into the structured observation space $\mathcal{O}_t$, represented in curvilinear coordinates $(s,d)$ around a given $\Gamma$.}
  \label{fig:obs_space}
\end{figure}
To integrate the observations into the WM, we apply separate encoders $\psi$. Ego-centric state vectors are processed by a multi-layer perceptron (MLP) $\bar{\psi}$, while both obstacle and lanelet sets are passed through permutation-invariant set autoencoders $\hat{\psi}$~\cite{Kortvelesy2023}. This yields encoded observations:
\begin{equation}
\begin{aligned}
\bar{z_t} = \bar{\psi}(\bar{o}^\mathrm{EC}_t) , \quad
\hat{z_t} = \bigl[\hat{\psi}(\hat{o}^\mathrm{obs}_t), \, \hat{\psi}(\hat{o}^{\mathrm{lane}}_t)\bigr]. 
\end{aligned}
\label{eq:encoder_concat}
\end{equation}
A key property of our implementation is that these set encoders are decodable. Approximate reconstructions of $\hat{o}^{\mathrm{obs}}_t$ and $\hat{o}^{\mathrm{lane}}_t$ must be recoverable from $s_t$, in order to prevent collapse into uninterpretable embeddings and provide a reconstruction signal essential for the WM training. While decodable sets have been proposed previously~\cite{Kortvelesy2023}, this is, to the best of our knowledge, the first integration of such a mechanism into Dreamer. Finally, the latent representation $z_t$ is conditioned on the encoded observations and recurrent state $h_t$:
\begin{equation}
z_t \sim q_{\varphi}(z_t \mid h_t, \bigl[ \bar{z_t}, \hat{z_t}\bigr]).
\label{eq:latent_distribution}
\end{equation}
%

The model states are refined by minimizing the weighted sum of the prediction, dynamics, and representation losses as in~\cite{Hafner2025}. The training objective is therefore:
\begin{equation}
\mathcal{L}_\mathrm{WM} = 
\beta_\mathrm{pred}\,\mathcal{L}_\mathrm{pred} +
\beta_\mathrm{dyn}\,\mathcal{L}_\mathrm{dyn} +
\beta_\mathrm{rep}\,\mathcal{L}_\mathrm{rep},
\label{eq:wm_loss}
\end{equation}
with loss weights $\beta_\mathrm{pred}$, $\beta_\mathrm{dyn}$, and $\beta_\mathrm{rep}$. Here, $\mathcal{L}_\mathrm{pred}$ is the prediction loss, $\mathcal{L}_\mathrm{dyn}$ the dynamics loss, and $\mathcal{L}_\mathrm{rep}$ regularizes the representation~\cite{Hafner2025}. These learned dynamics provide the critic $V_\phi$ with imagined rollouts in latent space, from which it estimates value distributions that guide the actor policy $\pi_\theta$.


\subsection{Reinforcement Learning Agent}
\label{subsec:reinforcementLearningAgent}
The RL agent operates on the WM states $s_t$ and outputs high-level goal specifications $g(\tau)$ as actions.
Formally, the interaction between the RL agent and the environment is modeled as a Markov decision process (MDP) with model state $s_t$, action $g(\tau)$, transition dynamics given by the WM, and scalar reward $r_t$. The output RL policy is denoted as:
\begin{equation}
    g(\tau) \sim \pi_\theta\bigl(\,\cdot \,\big|\, s_t\bigr),
    \label{eq:rl_policy}
\end{equation}
with parameters $\theta$ optimized to maximize expected return. Training follows the Dreamer paradigm~\cite{Hafner2025}, in which actor and critic are updated solely on imagined trajectories generated from the latent dynamics. The critic predicts a distribution over discounted returns for each model state, while the actor tries to maximize the cumulative reward for all model states by choosing the optimal $g(\tau)$.
The agent's action space is defined in the curvilinear frame $(s,d)$ around $\Gamma$. Instead of producing full trajectories, the agent selects terminal goal specifications $g(\tau)$ consisting of a lateral displacement $d$, a longitudinal terminal velocity $\dot s$, and a time horizon $\tau$. The time horizon $\tau$ specifies when the terminal state $(d, \dot s)$ should be reached and thereby controls the aggressiveness of the maneuver. The only fixed parameter in $g(\tau)$ is the terminal lateral velocity $\dot d$, which is set to zero to ensure parallel alignment with the reference path $\Gamma$ at the end of each generated trajectory. \autoref{tab:actions} summarizes the domains of all components of $g(\tau)$.

\begin{table}[!ht]
\centering
\renewcommand{\arraystretch}{1.1}
\caption{Domains of the goal state components $g(\tau)$.}
\begin{tabularx}{0.55\linewidth}{c c c}
    \toprule
    Symbol & Type & Domain \\
    \midrule
    $d$ & action & $[-\SI{5.0}{\meter},\; \SI{5.0}{\meter}]$ \\
    $\dot s$ & action & $[\SI{0}{\meter\per\second},\; \SI{15}{\meter\per\second}]$ \\
    $\tau$ & action & $[\SI{0.3}{\second},\; \SI{1.5}{\second}]$ \\
    $\dot d$ & fixed & $\SI{0}{\meter\per\second}$ \\
    \bottomrule
\end{tabularx}
\label{tab:actions}
\end{table}

The reward function encourages the agent to select actions $g(\tau)$ that enable steady progress along $\Gamma$ while remaining collision-free and within road boundaries. Episodes terminate upon collision, boundary violation, or upon reaching the goal area, such that maximizing the reward corresponds to successfully bypassing obstacles and rejoining $\Gamma$ within the designated target region. The reward function is defined as:

\vspace{-3mm}
\begin{small}
\begin{equation}
r_t =
\underbrace{\Delta s_t \,(1 - |d_t|)^2}_{\text{dense reward}}
\;+\;
\underbrace{
\begin{cases}
\;\;1, & \text{if } s \in \{\text{Goal}\}, \\
-1, & \text{if } s \in \{\text{Collision}, \text{Error}\}, \\
\;\;0, & \text{otherwise}.
\end{cases}
}_{\text{sparse reward}}
\end{equation}
\end{small}
The dense reward is defined based on incremental longitudinal progress along $\Gamma$ and lateral deviation from it. Let $\Delta s_t$ denote the normalized progress between two consecutive simulation steps, such that the cumulative sum of $\Delta s_t$ reaches $1$ when the goal area is reached, and let $d_t \in [-1,1]$ denote the normalized lateral deviation from $\Gamma$. The dense component provides continuous feedback for forward motion along $\Gamma$ while penalizing lateral deviation. If the ego vehicle overshoots the goal area without terminating, $\Delta s_t$ becomes negative and the sparse reward is not obtained, resulting in a decreasing accumulated return that discourages overshooting. The sparse reward explicitly encodes task completion and failure by assigning a terminal reward upon reaching the goal area or a penalty for collisions or boundary violations. We also evaluated dense-only and sparse-only reward formulations and found that their combination yields the most stable learning behavior and best overall performance.

\subsection{Deterministic Trajectory Generation \& Evaluation}
\label{subsec:deterministic_planner}
A deterministic motion planner analytically generates candidate trajectories from the initial condition \(x(0)\) to the goal specification \(g(\tau)\) provided by the RL agent, in the curvilinear frame \((s,d)\). A candidate trajectory \(\xi:[0,\tau]\!\to\!\mathbb{R}^m\) satisfies \(x(t)=\xi(t)\), with \(x(0)=\xi(0)\) and a terminal state \(x(\tau)=\xi(\tau)\) that is consistent with the degrees of freedom constrained by \(g(\tau)\). Formally, the planner returns:
\begin{equation}
    \xi \;=\; \mathcal{P}\bigl(x(0),\,g(\tau),\,\Gamma\bigr).
    \label{eq:planner_mapping}
\end{equation}
To realize the polynomial trajectory generation operator \(\mathcal{P}(\cdot)\), lateral and longitudinal subproblems are solved using jerk-optimal polynomial primitives~\cite{Werling2010}: a quintic polynomial for the lateral component \(d(t)\) and a quartic polynomial for the longitudinal component \(s(t)\), with boundary conditions derived from \(\{x(0),g(\tau)\}\). The polynomials are evaluated in the curvilinear frame, and the resulting trajectories are then transformed into the Cartesian frame using the curvilinear mapping described in~\cite{Werling2010}.

The planner forms a local set of candidate trajectories around the goal proposal \(g(\tau)\) returned by the RL agent.
In particular, the planner samples end points in a neighborhood \(\mathcal{G}\bigl(g(\tau)\bigr)\) of \(g(\tau)\), as illustrated in \autoref{fig:neighborhood_sampling}. The corresponding set of generated trajectories is:
\begin{equation}
    \mathcal{T} = \Bigl\{\, \xi_k =\mathcal{P}\bigl(x(0),\,\tilde g_k(\tau),\Gamma\bigr) \;\Big|\; \tilde g_k(\tau)\in \mathcal{G}\bigl(g(\tau)\bigr) \Bigr\}.
    \label{eq:candidate_set}
\end{equation}
%
\begin{figure}[!b]
  \centering
    \input{figures/sampling.tex}%

  \vspace{-4mm}
  \caption{Neighborhood sampling around the goal state $g(\tau)$ returned by the RL agent. The planner perturbs $g(\tau)$ within the local set $\mathcal{G}(g(\tau))$ and generates the corresponding candidate trajectories $\mathcal{T}$.}
  \label{fig:neighborhood_sampling}
\end{figure}

We first validate candidates with respect to the curvilinear domain of \(\Gamma\), where $\mathcal{D}_\Gamma$ is the projection domain defined along and around~$\Gamma$. Let \(\Phi_\Gamma:\mathcal{D}_\Gamma \subset \mathbb{R}^{2}\!\to\!\mathbb{R}^{2}\) denote the curvilinear-to-Cartesian projection, which is bijective on \(\mathcal{D}_\Gamma\) with inverse \(\Psi_\Gamma=\Phi_\Gamma^{-1}\). 
A candidate is valid if its curvilinear coordinates remain in \(\mathcal{D}_\Gamma\) for the entire horizon:
\begin{equation}
    \mathcal{T}_{\mathrm{valid}}
    \;=\;
    \bigl\{\, \xi \in \mathcal{T} \;\Big|\; \bigl(s_\xi(t),d_\xi(t)\bigr)\in \mathcal{D}_\Gamma,\ \forall t\in[0,\tau] \Bigr\}
    \label{eq:valid_subset}
\end{equation}
Kinematic and dynamic feasibility is then assessed in Cartesian coordinates, 
since only in this representation the road curvature is correctly taken into account. 
Bounds are checked on acceleration, curvature, curvature rate, and yaw rate:
\begin{equation}
\begin{aligned}
    \|a_\xi(t)\| &\le a_{\max}, 
    & |\kappa_\xi(t)| &\le \kappa_{\max}, \\
    |\dot{\kappa}_\xi(t)| &\le \dot{\kappa}_{\max},
    & |\dot{\psi}_\xi(t)| &\le \dot{\psi}_{\max}.
\end{aligned}
\label{eq:feas_bounds}
\end{equation}
We collect the scalar feasibility conditions from \eqref{eq:feas_bounds} into the family $\mathcal{H}_{\mathrm{kin}}$. 
Each $h \in \mathcal{H}_{\mathrm{kin}}$ abstracts one of these constraints, and $h(\xi)\le 0$ indicates that the corresponding condition is satisfied. The feasible set is therefore:
\begin{equation}
    \mathcal{T}_{\mathrm{feas}} = \{\xi \in \mathcal{T}_{\mathrm{valid}} \mid h(\xi)\!\le\!0\ \forall h\!\in\!\mathcal{H}_{\mathrm{kin}}\}.
    \label{eq:feasible_set}
\end{equation}
Costs are then evaluated using the weighted sum $J_\mathrm{sum}$ of all cost functions~\cite{Frenetix}. This allows task-specific terms such as comfort and safety. The selected trajectory is:
\begin{equation}
    \xi^\star \;=\; \arg\min_{\xi \in \mathcal{T}_\mathrm{feas}} J_\mathrm{sum}\!\bigl(\xi\bigr),
    \label{eq:planner_selection}
\end{equation}
and is passed to the controller or simulator. Optimality is determined by the analytic evaluation, meaning that the final choice does not depend on whether the resulting trajectory coincides with the one originally suggested by the RL agent.

\section{Results \& Discussion}
\label{sec:results}

We now report the empirical evaluation of our proposed hybrid motion planner. The objectives are to quantify planning performance and computational cost across a large set of scenarios, to qualitatively analyze representative cases, and to assess generalization to layouts outside the training distribution. All experiments are conducted in the CommonRoad (CR) simulation environment~\cite{Commonroad} using a workstation equipped with an AMD Ryzen 9 7950X CPU and an NVIDIA RTX 4090 GPU. The agent is trained for $1\mathrm{M}$ steps using DreamerV3~\cite{Hafner2025} with approximately $200\mathrm{M}$ parameters. No explicit hyperparameter tuning is performed. This follows prior work demonstrating that DreamerV3 can explore and achieve complex, long-term goals without extensive manual tuning~\cite{Hafner2025}. Our evaluation focuses on evasive maneuver scenarios in which the ego vehicle is forced to deviate from $\Gamma$ due to obstructing vehicles. This scenario class is intentionally chosen to assess the effectiveness of learned goal sampling. In unobstructed settings, feasible motion plans can often be obtained by merely adjusting longitudinal velocity along $\Gamma$, which does not require learned goal generation~\cite{Dauner2023CORL}. The evaluated scenarios are derived from the CR benchmark suite and are extended with generated instances. In these scenarios, between one and four vehicles are placed on one- or two-lane road segments, requiring the ego vehicle to perform an avoidance maneuver. Start and goal states are randomized and include lane changes where feasible. Test scenarios are selected such that they are not part of the training set. An example scenario is illustrated in \autoref{subsec:qualitative_evaluation}.

\subsection{Quantitative Results}
\label{subsec:quantitative_evaluation}

We benchmark RL-guided and classical sampling strategies on 300 evasive scenarios. Each setting defines how the candidate set $\mathcal{T}$ is constructed from the agent’s terminal specification $g(\tau)$. The setting \textbf{RL} evaluates only the goal proposal $g(\tau)$ generated by the agent. The hybrid settings \textbf{RL1}–\textbf{RL3} extend this proposal by additionally sampling goal candidates in local neighborhoods around $g(\tau)$, following the scheme illustrated in \autoref{fig:neighborhood_sampling}. Specifically, \textbf{RL1}, \textbf{RL2}, and \textbf{RL3} add 1, 2, and 3 neighborhood layers, respectively. Each successive layer increases the magnitude of the offsets $\Delta \dot{s}$ and $\Delta d$, thereby expanding the sampling region around $g(\tau)$. As a result, an increasing number of additional trajectories is generated and evaluated using~\eqref{eq:planner_selection}. As classical baselines~\cite{Frenetix}, we consider uniform terminal sampling with 800 candidates (\textbf{B800}) and a reduced variant with 125 candidates (\textbf{B125}). A single-sample baseline (\textbf{B1}) and a random sampler selecting $g(\tau)$ randomly from the action space were also evaluated. As both failed in all scenarios, they are omitted from further analysis.

We first analyze the quality of the generated trajectories across the evaluation suite. \autoref{fig:traj_quality_overall} summarizes success rates together with the occurrence of different failure modes.%
\begin{figure}[!ht]
  \centering
    \newcommand{\plotheight}{2.6cm}
\newcommand{\plotwidth}{6.6cm}

\pgfplotstableread{
Label	Success	Collision	Timelimit	Feasibility
base800	100	0	0	0
base125	99. 0.666 0.333333333	0
RL0	96	2	0	2
RL1	98	1.333333333	0	0.666666667
RL2	99	1	0	0
RL3	100	0	0	0
}\resultdatafinal

\begin{tikzpicture}[font=\footnotesize]
\begin{axis}[
    width  = \plotwidth,
    height = \plotheight,
    major x tick style = transparent,
    ybar stacked,
    bar width=15pt,
    ymajorgrids = true,
    ylabel = {Percentage in $\si{\percent}$},
    xtick = data,
    xticklabels = {B800, B125, RL, RL1, RL 2, RL3},
    ylabel shift={-6pt},
    scale only axis,
    scaled y ticks = false,
    enlarge x limits=0.15,
    axis y line*=left,
    ymin=80,
    ymax=100,
    legend cell align=left,
    legend style={
        at={(0.5,-0.17)}, 
        anchor=north,
        legend columns=2,
        cells={anchor=center},
        draw=none,
        column sep=1em,
        row sep=0.05em,
    },
]
  \addplot[style={Black,fill=Blue,mark=none}] table [y=Success, meta=Label, x expr=\coordindex] {\resultdatafinal};
  \addlegendentry{Success}

  \addplot[style={Black,fill=Orange,mark=none}] table [y=Collision, meta=Label, x expr=\coordindex] {\resultdatafinal};
  \addlegendentry{Collision}

  \addplot[style={Black,fill=LightGray,mark=none}] table [y=Feasibility, meta=Label, x expr=\coordindex] {\resultdatafinal};
  \addlegendentry{Kinematic Infeasibility}
  
  \addplot[style={Black,fill=Green,mark=none}] table [y=Timelimit, meta=Label, x expr=\coordindex] {\resultdatafinal};
  \addlegendentry{Timelimit}
  
\end{axis}


\begin{axis}[
/pgf/number format/.cd,
1000 sep={},
height=\plotheight,
width=\plotwidth,
enlarge x limits=0.15,
scale only axis,
scaled ticks=false,
scaled ticks=false,
ylabel={Sampled trajectories},
xmin=0,
xmax=1,
ymin=0, 
ymax=800,
axis x line*=none,
axis y line*=right,
ylabel shift={-4pt},
xticklabels={},
xtick={-10}
]

\addplot [thick, Black, mark=x, mark size=3pt, only marks]
table {%
0.00 800 
0.20 125  
0.40 1   
0.60 27   
0.80 125
1.00 343  
};

\end{axis}

\end{tikzpicture}%

  \vspace{-0.8cm}
  \caption{Aggregate performance across all evaluation scenarios. Bars indicate success and failure outcomes, while crosses (\ding{53}) denote the number of sampled trajectories. Although \textbf{B125} and \textbf{RL2} achieve similar success rates, our RL-guided variants concentrate sampling in promising regions, resulting in a higher fraction of feasible and drivable candidates per sample (\autoref{fig:traj_distribution}).}
  \label{fig:traj_quality_overall}
\end{figure}
The uniform baseline (\textbf{B800}) solves all scenarios, representing the performance of a dense but non-adaptive sampler. Reducing the number of candidates to \textbf{B125} still yields a high success rate of \SI{99}{\percent}. However, this coverage is achieved through uniform exploration of the action space, leading to a relatively sparse distribution of feasible trajectories within the relevant maneuver region. In contrast, our RL-guided variants concentrate sampling around promising goal proposals. Even without neighborhood refinement, \textbf{RL} achieves \SI{96}{\percent} success with just one trajectory. Adding neighborhood sampling improves robustness, and \textbf{RL1--RL3} achieve up to \SI{100}{\percent}.
Importantly, although \textbf{B125} and \textbf{RL2} achieve similar success rates with comparable numbers of sampled trajectories, their candidate distributions differ substantially. The uniform baseline generates many infeasible trajectories (\autoref{fig:traj_distribution}), whereas our RL-guided variants concentrate samples in the relevant maneuver region, resulting in a denser and more structured candidate set. To quantify this effect, \autoref{fig:traj_distribution} analyzes the share of feasible versus infeasible trajectories within the candidate set $\mathcal{T}$. All results are aggregated over the full evaluation suite and all generated trajectories. The uniform baseline is reported in aggregate form, since its distribution remains nearly identical across different sample counts.
\begin{figure}[!ht]
  \centering
    \begin{tikzpicture}[font=\footnotesize]
    \begin{axis}[
        width  = 7cm,
        height = 2.8cm,
        major x tick style = transparent,
        ybar=0.5mm,
        bar width=6pt,
        ymajorgrids = true,
        ylabel = {Percentage in $\si{\percent}$},
        symbolic x coords={Base, RL, RL1, RL2, RL3},
        xtick = data,
        xticklabels = {Base, RL, RL1, RL2, RL3},
        scale only axis,
        scaled y ticks = false,
        enlarge x limits=0.15,
        ymin=0,
        ymax=80,
        ytick = {0, 20, 40, 60, 80},
        yticklabels = {0, 20, 40, 80, 100},
        y filter/.code={
          \pgfmathparse{
            (#1 <= 80) * (#1) +
            (#1 > 80)  * (#1 * 0.8)
          }
          \pgfmathresult
        },
        legend cell align=left,
        legend style={
        	at={(0.5,-0.19)}, 
        	anchor=north,
        	legend columns=2,
        	cells={anchor=center},
        	draw=none,
        	column sep=1em,
        	row sep=0.1em,
        },
        after end axis/.code={
            \def\breaky{0.625}
            \def\lw{3pt}
            \def\yoffset{0.015}
            \def\xoffset{0.02}
            \foreach \x in {0.059, 0.252, 0.444, 0.636, 0.829}{
            \draw [line width=\lw, white, decorate, decoration={snake, amplitude=1pt}] (rel axis cs:\x-\xoffset, \breaky) -- (rel axis cs:\x+\xoffset, \breaky);}
            \draw [line width=3pt, white] (rel axis cs:-0.02,\breaky) -- (rel axis cs:0.02,\breaky);
            \draw [line width=0.6pt, black] (rel axis cs:-0.02,\breaky+\yoffset) -- (rel axis cs:0.02,\breaky+\yoffset);
            \draw [line width=0.6pt, black] (rel axis cs:-0.02,\breaky-\yoffset) -- (rel axis cs:0.02,\breaky-\yoffset);
            \draw [line width=3pt, white] (rel axis cs:0.98,\breaky) -- (rel axis cs:1.02,\breaky);
            \draw [line width=0.6pt, black] (rel axis cs:0.98,\breaky+\yoffset) -- (rel axis cs:1.02,\breaky+\yoffset);
            \draw [line width=0.6pt, black] (rel axis cs:0.98,\breaky-\yoffset) -- (rel axis cs:1.02,\breaky-\yoffset);
        }
    ]
    \addplot[style={Black,fill=Blue,mark=none}]
        coordinates {(Base, 47.05) (RL, 95.91) (RL1, 93.44) (RL2, 88.74) (RL3, 82.93)};
    \addlegendentry{Feasible}

    \addplot[style={Black,fill=Orange,mark=none}]
         coordinates {(Base, 3.82) (RL, 0.65) (RL1, 0.85) (RL2, 1.36) (RL3, 1.85)};
    \addlegendentry{Obstacle Collision}

    \addplot[style={Black,fill=Bluelight,mark=none}]
         coordinates {(Base, 31.29) (RL, 3.35) (RL1, 5.56) (RL2, 9.60) (RL3, 14.56)};
    \addlegendentry{Road Boundary Collision}
         
    \addplot[style={Black,fill=LightGray,mark=none}]
        coordinates {(Base, 17.83) (RL, 0.10) (RL1, 0.15) (RL2, 0.30) (RL3, 0.66)};
    \addlegendentry{Kinematic Infeasibility}
    \end{axis}
\end{tikzpicture}%

  \vspace{-3mm}
  \caption{Distribution of trajectory outcomes aggregated over all evaluation scenarios. The baseline corresponds to uniform sampling (\textbf{B800}/\textbf{B125}), whose outcome distribution is nearly identical for different sample counts. Our RL-guided variants increase the proportion of feasible trajectories compared to uniform sampling, demonstrating more effective allocation of samples in the action space.}
  \label{fig:traj_distribution}
\end{figure}
The baselines yield only \SI{47}{\percent} feasible trajectories, with the remainder lost to collisions (\SI{35}{\percent}) or kinematic infeasibility (\SI{18}{\percent}). In contrast, our RL-guided planner places samples much more effectively: \textbf{RL} achieves \SI{96}{\percent} feasibility within its neighborhood, while the hybrid setting \textbf{RL3} achieves \SI{83}{\percent}. This demonstrates that the learned policy directs sampling toward meaningful regions of the action space.

The minimum distance to obstacles provides an additional perspective on trajectory quality. \autoref{fig:min_dist} summarizes the distributions across all evaluation scenarios. 
\begin{figure}[!b]
  \centering
    \begin{tikzpicture}[font=\footnotesize]
  \begin{axis}[
    width=7.5cm,
    height=2.8cm,
    scale only axis,
    scaled ticks=false,
    scaled ticks=false,
    xmin=0,
    xmax=2.5,
    boxplot/draw direction=x,
    xlabel={Minimum distance to obstacles in \si{\meter}},
    yticklabels={RL3, RL2, RL1, RL, B125, B800},
    ytick={1, 2, 3, 4, 5, 6},
    boxplot/box extend=0.7,
    ymajorgrids = true,
  ]
    \addplot[
      mark=*,
      mark options={color=black, scale=0.7},
      fill=Bluelight,
      boxplot prepared={
        lower whisker=0.697117,
        lower quartile=0.835488,
        median=1.041735,
        upper quartile=1.727345,
        upper whisker=1.870476,
        average=1.245262
      },
      draw=black
    ] coordinates {
    };
    \addplot[
      mark=*,
      mark options={color=black, scale=0.7},
      fill=Bluelight,
      boxplot prepared={
        lower whisker=0.719806,
        lower quartile=0.841621,
        median=1.093333,
        upper quartile=1.555304,
        upper whisker=1.842115,
        average=1.193509
      },
      draw=black
    ] coordinates {
    };
    \addplot[
      mark=*,
      mark options={color=black, scale=0.7},
      fill=Bluelight,
      boxplot prepared={
        lower whisker=0.574025,
        lower quartile=0.779903,
        median=1.089078,
        upper quartile=1.310396,
        upper whisker=1.865009,
        average=1.098663
      },
      draw=black
    ] coordinates {
    };

    \addplot[
      mark=*,
      mark options={color=black, scale=0.7},
      fill=Blue,
      boxplot prepared={
        lower whisker=0.067291,
        lower quartile=0.605471,
        median=0.900525,
        upper quartile=1.263404,
        upper whisker=1.555127,
        average=0.914079
      },
      draw=black
    ] coordinates {
    };

    \addplot[
      mark=*,
      mark options={color=black, scale=0.7},
      fill=LightGray,
      boxplot prepared={
        lower whisker=0.754811,
        lower quartile=0.906034,
        median=1.132825,
        upper quartile=1.739234,
        upper whisker=2.008118,
        average=1.312274
      },
      draw=black
    ] coordinates {
    };
    
    \addplot[
      mark=*,
      mark options={color=black, scale=0.7},
      fill=LightGray,
      boxplot prepared={
        lower whisker=0.712940,
        lower quartile=0.916263,
        median=1.059183,
        upper quartile=1.794715,
        upper whisker=1.795867,
        average=1.266755
      },
      draw=black
    ] coordinates {
    };
  \end{axis}
\end{tikzpicture}%

  \vspace{-4mm}
  \caption{Distribution of minimum obstacle distances across all evaluation scenarios. While the pure \textbf{RL} proposal can yield smaller clearances due to the absence of explicit distance shaping during training, the hybrid variants (\textbf{RL1--RL3}) maintain safety margins comparable to those of the uniform baselines through neighborhood sampling.}
  \label{fig:min_dist}
\end{figure}
We stress that the RL agent has not been trained to explicitly maximize safety margins. For the policy, the only relevant signal is to follow the reference path to the goal area and thereby indirectly to avoid collisions. While \textbf{RL} yields feasible trajectories, their clearance to obstacles can be comparatively small. The \textbf{RL1--RL3} variants achieve larger average safety margins, as the analytic evaluation explicitly rewards maintaining a safe distance from obstacles.

A further important aspect concerns computation time. In dense traffic, classical approaches often bias sampling toward obstacle-rich regions to increase the chance of finding feasible motion, e.g.~\cite{Li2020 }. This strategy, however, becomes counterproductive, since the additional candidates must all be checked for feasibility and collisions, which dominates the runtime. The strength of our hybrid planner lies in focusing samples directly in promising regions. This advantage becomes apparent already with two obstacles. While \textbf{B800} spends considerable time discarding infeasible or irrelevant trajectories (\autoref{fig:traj_distribution}), our method evaluates only a fraction of candidates without loss of performance. \autoref{fig:traj_time} reports the runtime per planning cycle, including trajectory generation and evaluation. 
\begin{figure}[!ht]
  \centering
    \begin{tikzpicture}[font=\footnotesize]
  \begin{axis}[
    width=6.5cm,
    height=3.4cm,
    scale only axis,
    scaled ticks=false,
    xmin=0,
    xmax=22,
    boxplot/draw direction=x,
    xlabel={Runtime in \si{\milli\second}},
    yticklabels={
      B800 (\textcolor{Blue}{376}),
      B605 (\textcolor{Blue}{284}),
      B236 (\textcolor{Blue}{111}),
      B125 (\textcolor{Blue}{59}),
      RL (\textcolor{Blue}{1}),
      RL1 (\textcolor{Blue}{25}),
      RL2 (\textcolor{Blue}{111}),
      RL3 (\textcolor{Blue}{284})
    },
    ytick={8,7,6,5,4,3,2,1},
    boxplot/box extend=0.7,
    xlabel shift=-1mm,
    enlarge y limits=0.07,
    ymajorgrids = true,
  ]

    \addplot[
      fill=Bluelight,
      draw=black,
      boxplot prepared={
        lower whisker=5.013227,
        lower quartile=5.395472,
        median=5.688071,
        upper quartile=8.210480,
        upper whisker=12.427568,
        average=7.600401
      }
    ] coordinates {};

    \addplot[
      fill=Bluelight,
      draw=black,
      boxplot prepared={
        lower whisker=3.019571,
        lower quartile=3.257036,
        median=3.414392,
        upper quartile=4.364014,
        upper whisker=6.022692,
        average=4.266087
      }
    ] coordinates {};

    \addplot[
      fill=Bluelight,
      draw=black,
      boxplot prepared={
        lower whisker=2.045631,
        lower quartile=2.214670,
        median=2.301931,
        upper quartile=2.626657,
        upper whisker=3.244638,
        average=2.561150
      }
    ] coordinates {};

    \addplot[
      fill=Blue,
      draw=black,
      boxplot prepared={
        lower whisker=1.821041,
        lower quartile=1.981258,
        median=2.036810,
        upper quartile=2.106905,
        upper whisker=2.294540,
        average=2.064128
      }
    ] coordinates {};

    \addplot[
      fill=LightGray,
      draw=black,
      boxplot prepared={
        lower whisker=2.710104,
        lower quartile=3.280997,
        median=3.449678,
        upper quartile=4.178762,
        upper whisker=5.523205,
        average=4.025068
      }
    ] coordinates {};

    \addplot[
      fill=LightGray, 
      draw=black,
      boxplot prepared={
        lower whisker=5.2,
        lower quartile=5.8,
        median=6.2,
        upper quartile=7.3,
        upper whisker=10.5,
        average=6.9
      }
    ] coordinates {};

    \addplot[
      fill=LightGray, 
      draw=black,
      boxplot prepared={
        lower whisker=7.0,
        lower quartile=8.4,
        median=9.1,
        upper quartile=11.8,
        upper whisker=18.5,
        average=10.6
      }
    ] coordinates {};

    \addplot[
      fill=LightGray,
      draw=black,
      boxplot prepared={
        lower whisker=6.903648,
        lower quartile=8.751631,
        median=9.447455,
        upper quartile=13.745785,
        upper whisker=21.224022,
        average=13.120452
      }
    ] coordinates {};

  \end{axis}
\end{tikzpicture}%

  \vspace{-4mm}
  \caption{Runtime per planning cycle across all evaluation scenarios. Blue numbers denote the average number of feasible trajectories per cycle.  Uniform baselines \textbf{B236} and \textbf{B605} are scaled to match the feasible trajectory counts of our \textbf{RL2} and \textbf{RL3}. Our RL-guided variants achieve the same number of feasible trajectories with significantly fewer total samples (\autoref{fig:traj_quality_overall}) and lower runtime, highlighting their improved sampling efficiency.}
  \label{fig:traj_time}
\end{figure}
Our \textbf{RL} setting is fastest at about \SI{2}{\milli\second}, since only one trajectory is generated. Runtime increases with neighborhood size, with \textbf{RL3} slightly exceeding \textbf{B125} in absolute cycle time. However, absolute runtime alone does not reflect sampling efficiency. The scaled baselines \textbf{B236} and \textbf{B605} correspond to uniform sampling with increased sample counts (236 and 605, respectively) such that the expected number of feasible trajectories matches those of our \textbf{RL2} (111 feasible) and \textbf{RL3} (284 feasible), respectively. This shows that, for the same number of feasible trajectories, our RL-guided variants require fewer samples and less computation time than uniform sampling. The WM adds a fixed overhead of \SI{3.75}{\milli\second}, which is not included in~\autoref{fig:traj_time}. This overhead is constant per cycle and independent of the number of sampled trajectories. Moreover, the WM representation enables future extensions such as reasoning over occluded regions.

\subsection{Qualitative Analysis}
\label{subsec:qualitative_evaluation}

To complement the evaluation, we consider a representative evasive maneuver in detail. In this scenario, the ego vehicle must bypass another vehicle to reach its goal. \autoref{fig:qualitative_evaluation} summarizes the analysis for the uniform baselines and our RL-guided variants.
\begin{figure*}[!ht]
  \centering
    \input{figures/qualitative_evaluation}%

  \vspace{-4mm}
  \caption{Qualitative evaluation of an evasive maneuver. Our RL-guided hybrid approaches concentrate trajectories in the relevant maneuver region, resulting in an equally effective evasive maneuver with significantly fewer evaluated candidates. The central plot shows the fraction of drivable trajectories (solid lines) and the lateral deviation $d$ (dashed lines) as a function of the longitudinal position $s$. The upper panels depict the sampled candidate sets.}
  \label{fig:qualitative_evaluation}
\end{figure*}
The central plot reports the fraction of drivable trajectories, i.e., feasible and collision-free candidates, as a function of the longitudinal $s$-position. In addition, the dashed curves show the lateral offset $d$ of the selected motion relative to $\Gamma$. Bird’s-eye views of the scene are placed above the plot for \textbf{B800} and for the RL-guided hybrid planner (\textbf{RL2}).

The contrast between the two strategies becomes apparent in this example. The baseline methods distribute samples widely across feasible and infeasible regions, which results in many trajectories being rejected by feasibility checks or collision detection. By comparison, our hybrid approach concentrates goal proposals $g(\tau)$ in the viable corridor, and neighborhood sampling increases the density of valid options in this region. The analytic evaluation then selects a smooth maneuver. Compared to the uniform baseline, the hybrid planner provides a more structured candidate set and reduces wasted effort on infeasible motion. 

The example also highlights qualitative differences in the spatial behavior. As discussed in \autoref{subsec:quantitative_evaluation}, a pure RL planner without neighborhood sampling tends to generate solutions that pass closer to obstacles. This effect is mitigated by local diversification, which shifts the selected trajectory towards safer clearance. While such failure cases are not visible in the present example, the enriched sets $\mathcal{T}$ generated by neighborhood sampling $\mathcal{G}(g(\tau))$ increase the likelihood of recovering a drivable trajectory whenever the initial RL proposal $g(\tau)$ lies close to a constraint boundary. If no valid trajectories remain, the feasible set $\mathcal{T}_\mathrm{feas}$ is empty, and the planner switches back to regular uniform sampling, which we count as a failure. Despite this limitation, the qualitative case demonstrates that RL-guided sampling increases the density of feasible trajectories. This provides the analytic planner with richer alternatives for selecting safe maneuvers and illustrates how learning-based guidance and analytic evaluation complement each other in practice.

\subsection{Generalization to Unseen Layouts}
We assess generalization beyond the training distribution by comparing our RL-guided planner with FISS+~\cite{Sun2023} in two complementary scenarios: a standard CR scenario (DEU\_Lohmar-65\_1\_T-1) originally used in FISS+ and an evasive maneuver case. Both are evaluated without retraining. In the CR scenario, FISS+ performs strongly, and our hybrid planner achieves comparable results when neighborhood sampling is enabled, indicating that the learned terminal specifications $g(\tau)$ generalize to unseen road geometries. In contrast, the evasive maneuver scenario reveals a limitation of FISS+. Here, avoiding the obstacle requires deviating from $\Gamma$, which FISS+ cannot produce. Instead, it reacts conservatively by decelerating and remaining behind the obstructing vehicle, as shown in \autoref{tab:fiss_comparison}. Overall, our planner handles both cases, while FISS+ remains domain-limited.
\begin{table}[!ht]
\centering
\renewcommand{\arraystretch}{1.1}
\caption{Comparison of FISS+ and our planner in two scenarios.}

\label{tab:fiss_comparison}
\begin{tabularx}{0.95\linewidth}{l c c c c c}
    \toprule
    Scenario & Method & $v(\SI{3}{\second})$ & $v(\SI{5}{\second})$ & $v(\SI{7}{\second})$ & Goal \\
    \midrule
    \multirow{2}{*}{FISS+} 
      & FISS+~\cite{Sun2023} & \SI{10.8}{\meter\per\second} & \SI{11.8}{\meter\per\second} & \SI{12.8}{\meter\per\second} & \checkmark \\
      & \textbf{Ours}  & \SI{13.7}{\meter\per\second} & \SI{14.6}{\meter\per\second} & \SI{14.9}{\meter\per\second} & \checkmark \\
    \midrule
    \multirow{2}{*}{Evasive test}    
      & FISS+~\cite{Sun2023} & \SI{9.0}{\meter\per\second} & \SI{7.2}{\meter\per\second}   &  \SI{5.0}{\meter\per\second}   & \ding{53} \\
      & \textbf{Ours } &  \SI{14.1}{\meter\per\second}  &  \SI{11.6}{\meter\per\second}  &  \SI{13.6}{\meter\per\second}  & \checkmark \\
    \bottomrule
\end{tabularx}
\end{table}

\section{Conclusion \& Outlook}
\label{sec:conclusion}

In this work, we presented a hybrid motion planning framework that combines WM-based RL for goal state sampling with analytic trajectory generation and evaluation to preserve interpretability. Our hybrid method outperforms uniform sampling across computation time, feasible maneuvers, and sample efficiency, while maintaining planning quality and demonstrating strong generalization. Qualitative case studies further illustrated how the agent concentrates candidates in viable corridors. Nevertheless, several limitations remain. So far, the agent has been trained and evaluated exclusively in CR scenarios and tailored evasive maneuver settings. This setup does not fully capture the complexity of real-world traffic. Extending training and evaluation to real-world datasets and more complex traffic scenarios is, therefore, a necessary next step. Moreover, the planner currently lacks a mature backup strategy when the feasible set $\mathcal{T}_\mathrm{feas}$ becomes empty. Future work will address these limitations and extend the framework. We aim to augment the agent with an uncertainty estimate, allowing it to specify how many additional trajectories to sample in $\mathcal{G}(g(\tau))$. Finally, we plan to implement and evaluate the algorithm on a real vehicle to validate its effectiveness under real-world conditions.


\bibliographystyle{IEEEtran}
\bibliography{literature}
\end{document}